# Title: Few-Shot Deployment of Pretrained MRI Transformers in Brain Imaging Tasks

## Authors/affiliations

Mengyu Li [1,2], Guoyao Shen [1,2], Chad W. Farris *[3], and Xin Zhang*[1,2,4,5,6]

[1] Department of Mechanical Engineering, Boston University, Boston, MA 02215, United States

[2] The Photonics Center, Boston University, Boston, MA 02215, United States

[3] Department of Radiology, Boston University Chobanian & Avedisian School of Medicine, Boston, MA 02118, United States

[4] Department of Electrical and Computer Engineering, Boston University, Boston, MA, United States

[5] Department of Biomedical Engineering, Boston University, Boston, MA, United States

[6] Division of Materials Science and Engineering, Boston University, Boston, MA, United States

**Co-author:** Guoyao Shen, email: guoyao@bu.edu

**\*Corresponding author:** Prof. Xin Zhang, email: xinz@bu.edu; Dr. Chad Farris, email: cfarris@bu.edu

## Abstract

Machine learning using transformers has shown great potential in medical imaging, but its real-world applicability remains limited due to the scarcity of annotated data. In this study, we propose a practical framework for the few-shot deployment of pretrained MRI transformers in diverse brain imaging tasks. By utilizing the Masked Autoencoder (MAE) pretraining strategy on a large-scale, multi-cohort brain MRI dataset comprising over 31 million slices, we obtain highly transferable latent representations that generalize well across tasks and datasets. For high-level tasks such as classification, a frozen MAE encoder combined with a lightweight linear head achieves state-of-the-art accuracy in MRI sequence identification with minimal supervision. For low-level tasks such as segmentation, we propose MAE-FUnet, a hybrid architecture that fuses multiscale CNN features with pretrained MAE embeddings. This model consistently outperforms other strong baselines in both skull stripping and multi-class anatomical segmentation under data-limited conditions. With extensive quantitative and qualitative evaluations, our framework demonstrates efficiency, stability, and scalability, suggesting its suitability for low-resource clinical environments and broader neuroimaging applications.

# 1. Introduction

Recent advances in deep learning have made remarkable progress in medical imaging interpretation [1-5]. A variety of machine learning architectures have been widely investigated for medical analysis, including CNNs [6, 7], MLPs [8], transformers [9-11], and diffusion models [12-14]. Transformer-based architectures – initially proposed from the NLP community [15-18] – demonstrate strong generalization across vision and multi-modal tasks, which has been leveraged for various medical tasks [19, 20]. Furthermore, transformers have shown particular effectiveness in tasks such as brain structure segmentation, lesion detection, and image classification [21-23]. Their long-range cross-attention mechanisms allow them to encode global spatial context much more effectively than conventional CNNs or MLPs. However, transformers' large number of trainable parameters and attention-based structure often result in high data consumption. Thus, adapting them to real-world medical conditions is still challenging due to the limited availability of labeled samples.

In clinical neuroimaging, data scarcity remains a common obstacle due to privacy restrictions, annotation cost, and heterogeneity across imaging protocols. Most popular strategies that attempt to address this scenario typically fall into one of several categories: meta-learning approaches that involve complex fine-tuning procedures [24–27], methods restricted to specific downstream tasks [28, 29], techniques limited to well-established 3D MRI datasets [30], or frameworks that rely on manual input [31]. Fortunately, these limitations can be addressed by leveraging vision transformers pretrained on large-scale, universal 2D datasets, which can be effectively deployed to various downstream tasks under few-shot conditions. One of the most widely used pretraining methods for vision transformers is the Masked Autoencoder (MAE) [32], which initiates pretraining by reconstructing the masked patches in 2D images. In our work, we focus on utilizing a similar MAE pretraining strategy on a large 2D brain MRI cohort to obtain a pretrained vision transformer architecture as a strong initialization. Unlike the task-specific models, pretrained vision transformers can be repurposed with appropriate fine-tuning designs, offering promising performance even when trained under small-sample regimes. We systematically evaluate the MAE-based, few-shot architectures on non-pathological brain MRI tasks, including both classification and segmentation across diverse datasets.

Our approach targets multiple key benchmarks. Given the nature of different tasks, we construct tailored model variants to explore the most effective ways to utilize the learned embedding from the MAE baseline. For classification, we investigate the performance of a lightweight MAE-based classifier in discriminating MRI sequence modalities such as T1, T2, Fluid-Attenuated Inversion Recovery (FLAIR), Photo Density (PD), diffusion imaging (Diffusion Tensor Imaging (DTI) and Diffusion Weighted Imaging (DWI)). For segmentation, we assess an MAE variant that fuses the transformer's latent space embeddings with the U-Net features across multiple layers. This architecture, referred to as MAE-Fused U-Net (MAE-FUnet), is rigorously tested on both skull stripping and multi-class anatomical segmentation across multiple datasets, including Neurofeedback Skull-stripped (NFBS) [33], SynthStrip [34], and MRBrainS18 [35]. For each task domain, the MAE-based variants are compared against multiple well-established baselines—EfficientNetV2 [36], ResNet [37], and MedViT [38] for classification, and Swin-Unet [39], U-Net [6], TransUNet [9], and Segformer [40] for segmentation. Results consistently show that pretrained MAE models achieve superior or comparable accuracy in few-shot scenarios.

This study highlights the practicality and scalability of deploying pretrained transformers in real-world, data-limited brain imaging pipelines. Our findings support the adaptation of generalized MAE-based architectures into specialized variants tailored for classification and segmentation, demonstrating the possibility of reutilizing a generalized learned embedding on different task modalities. By emphasizing the few-shot transferability, this work also contributes to the customization of high-performance AI tools for neuroimaging, especially in clinical environments when targeting specific tasks with limited data resources.

## 2. Methods

### 2.1 MAE pretraining

The MAE pretraining implemented in this study builds on the original Masked Autoencoder framework by adjusting it specifically for brain MRI datasets. Similar to the original MAE, the model consists of a large transformer encoder and a lightweight decoder, with the target set to reconstruct masked image patches. The detail of the model structure is demonstrated in Figure 1a. The original MAE computes the reconstruction loss over the masked patches, while our loss function also introduces a sample-wise weighting scheme to address the partial-brain coverage that commonly occurs in aggregated MRI datasets. Using the brain masks generated from FreeSurfer (available http://surfer.nmr.mgh.harvard.edu/) as reference [41-54], we ensure that scans with limited brain content—such as those dominated by non-brain tissue or cropped fields of view—are appropriately down-weighted during training. The pipeline for sample-specific weighted loss is illustrated in Figure 1b. For the whole batch with $N$ samples, we compute the final loss as the weighted average of the per-sample losses $\mathcal{L}$:

$$\mathcal{L} = \frac{1}{N}\sum_{i=1}^{N} w_i \tilde{l}_i \tag{1}$$

where $\tilde{l}_i$ is the per-sample loss, averaged over masked patches; $w_i$ is a sample-specific weighting factor based on the brain area coverage of the sample $i$. The detailed mathematical work and training implementation can be found in Supplementary Information S.1.

To invoke the generalizability of the MAE transformer structure, a large pretraining dataset is required. We constructed our dataset from five different cohorts: National Alzheimer's Coordinating Center (NACC), Open Access Series of Imaging Studies (OASIS) [55-58], Alzheimer's Disease Neuroimaging Initiative (ADNI), FastMRI [59], and RadImageNet [60]. Each consists of a large amount of brain MRI with sequences spanning T1, T2, FLAIR, PD, and DWI. With the exclusion of other MRI regions, such as knee MRI in FastMRI, the final dataset is a large, comprehensive brain MRI dataset with 31 million 2D slices in total. All 2D slices are preprocessed into the shape of 224x224 with a 0.1% to 99.9% pixel value clamp and max-min normalization. Data argumentation includes random rotation, flipping, and cropping. The pretrained model size is set to be the same as ViT-Base from the original MAE paper, with a 16x16 patch size and 12 layers of transformer modules in the encoder.

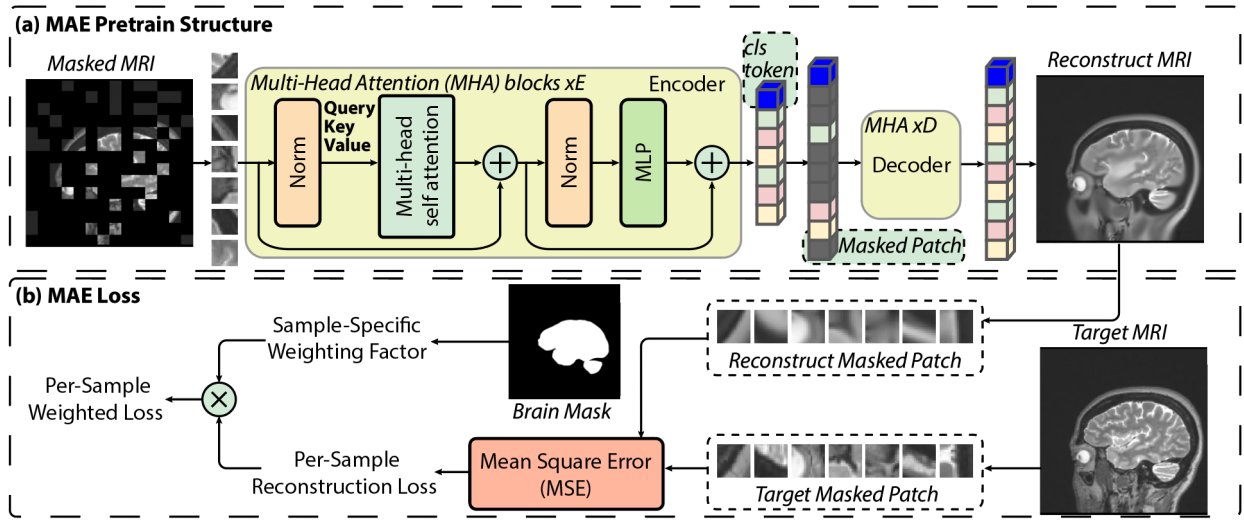

**Figure 1.** (a) Overview of the proposed MAE pretraining structure for brain MRI. MRI patches are encoded via a transformer encoder with $E$ Multi-Head Attention (MHA) blocks, followed by a lightweight decoder ($D$ MHA blocks) that reconstructs only the masked regions. (b) Illustration of the sample-specific loss computation. Mean squared error is calculated on reconstructed masked patches, and the per-sample loss is adaptively weighted based on brain-region coverage derived from a binary brain mask.

### 2.2 Direct Classification

To assess the utilization of pretrained transformer encoders in MAE for straightforward classification tasks such as sequence detection, we implemented a direct classification strategy using only the pretrained classification [CLS] token. Our approach repurposes the MAE transformer as a fixed feature extractor when adapting to downstream tasks. During fine-tuning, the encoder weights are frozen to preserve the learned representations, leading to fast downstream adaptations with fewer trainable parameters.

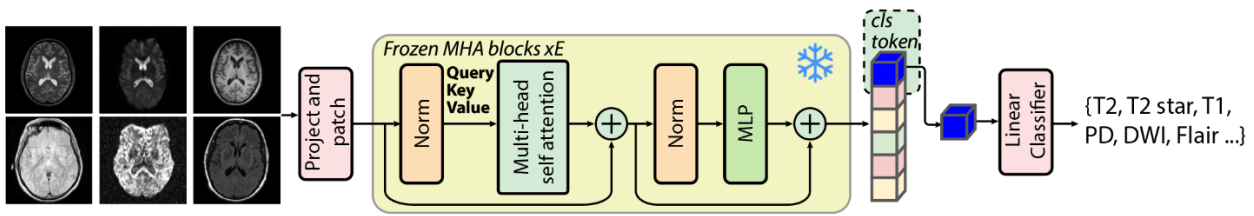

**Figure 2.** Direct classification pipeline for MRI sequence detection. Input MRI slices from multiple sequences are first tokenized via patch embedding and passed through the pretrained MAE encoder, consisting of $E$ frozen multi-head attention blocks. A [CLS] token summarizes the global representation, which is then fed into a lightweight linear classifier to predict the MRI sequence label (e.g., T1, T2, FLAIR, DWI, PD, T2*).

As illustrated in Figure 2, using sequence detection tasks as an example, input MRI slices from various image modalities, such as T1, T2, FLAIR, DWI, PD, and T2*, are first tokenized using a patch embedding projection layer. Then the patched tokens are processed through the frozen MAE encoder, comprising multiple stacked multi-head attention (MHA) and feedforward layers. The final [CLS] token embedding encodes the global context of the input image and is used as input to a lightweight linear classifier. The classifier is trained to predict the MRI sequence modality label in a supervised fashion using the AdamW optimizer [61] and $1 \times 10^{-4}$ learning rate. For multi-class logits classification task, such as sequence detection, the loss function is set to the cross-entropy loss:

$$\mathcal{L}_{ce} = -\log\left(\frac{e^{z_y}}{\sum_{j=1}^{C} e^{z_j}}\right) = -z_y + \log\left(\sum_{j=1}^{C} e^{z_j}\right) \qquad (2)$$

Where $z_j$ is the raw logit for class $j$, $y$ is the correct class index, and $C$ is the number of classes.

This linear probing setup serves two critical purposes: (1) evaluating the discriminative quality of the learned representations from MAE pretraining, and (2) demonstrating the few-shot adaptability of the frozen MAE transformers to classification tasks with minimal parameter overhead. Since the inputs are random, raw MRI slices with no modality-specific augmentation or pre-filtering, this pipeline is broadly applicable to heterogeneous real-world MRI datasets.

### 2.3 Segmentation with Fused Embedding

#### 2.3.1 MAE Fusion Architecture

To effectively adapt the pretrained MAE backbone for pixel-level tasks such as skull stripping and anatomical segmentation, we design a hybrid encoder-decoder architecture that fuses hierarchical CNN features with pretrained MAE transformer embeddings. This MAE-fused U-Net structure is referred to as the *MAE-FUnet*. As shown in Figure 3a, input MRI slices are first tokenized into patches and passed through the pretrained MAE transformer with frozen weights. In parallel, the same images are processed through a CNN-based U-Net backbone to extract multiscale spatial features. To integrate global representation from the transformer layers with local CNN features, we introduced a series of fusion blocks throughout multiple decoder levels. These fusion operations aim to create effective incorporation between local texture patterns (via CNNs) and global contextual information (via MAE).

The primary motivation for introducing additional structures, such as MAE-FUnet for pixel-wise downstream tasks, is to explore the full utilization of the rich contextual representations learned by MAE across multiple hierarchical layers. By incorporating both shallow and deep features, we aim to enable a more comprehensive and effective utilization of the MAE-derived embeddings. Compared to a direct deployment of CNN-based segmentation heads appended directly to the pretrained MAE output (analogous to the direct classification setup in 2.2), the fused architecture, MAE-FUnet, yields significantly improved segmentation performance. Detailed comparative evaluations are provided in the Results section.

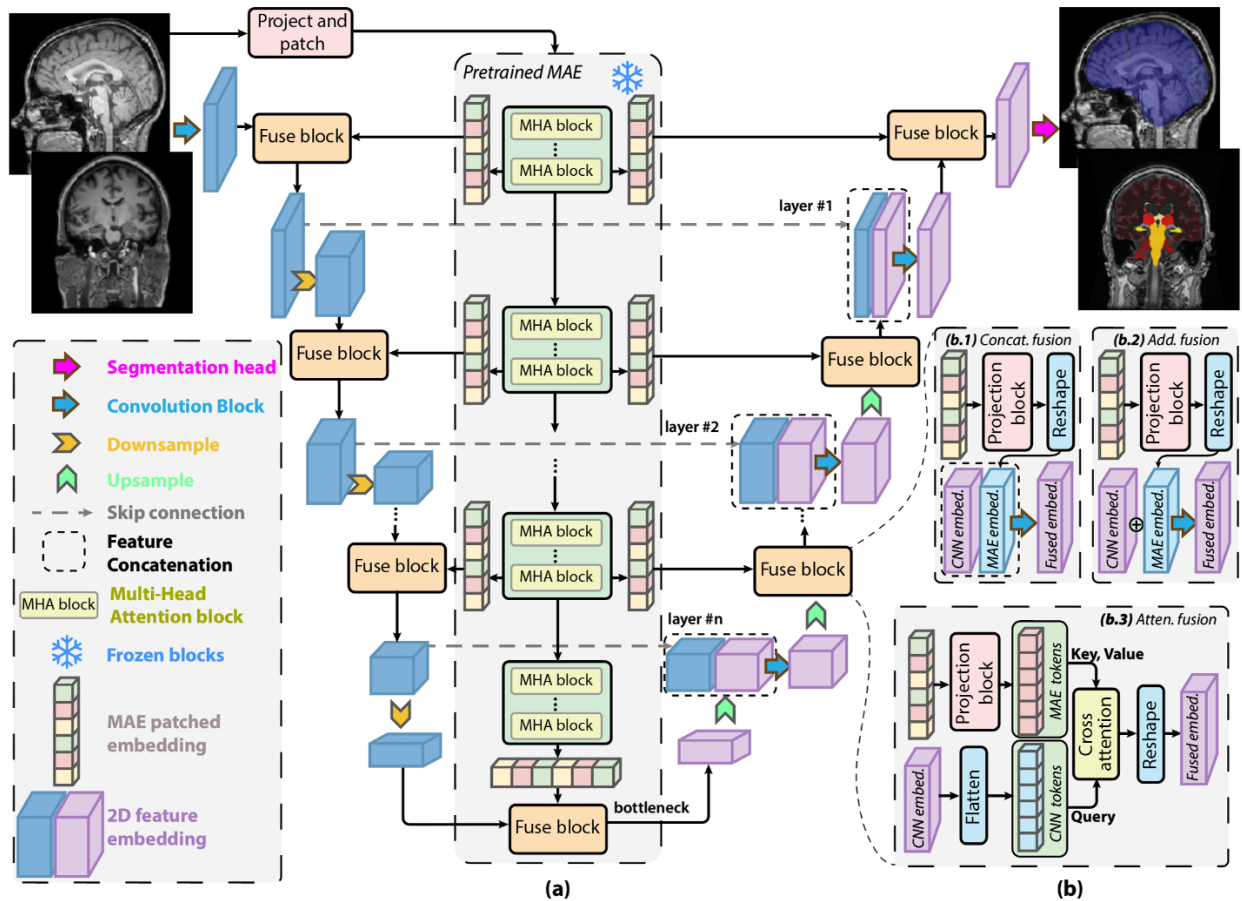

**Figure 3.** MAE-FUnet architecture for multi-class brain segmentation. (a) Overall structure fuses hierarchical CNN features with frozen MAE transformer embeddings through progressive decoder stages, incorporating skip connections and multi-scale feature integration. (b) Three possible fusion strategies are explored: (b.1) concatenation-based fusion, (b.2) addition-based fusion, and (b.3) cross-attention-based fusion between CNN and transformer features.

Three distinct fusion strategies are explored — concatenation, addition, and cross-attention — illustrated in Figure 3b:

**(b.1) Concatenation fusion:** Features from CNN and MAE streams are concatenated channel-wise after projection and reshaping.
**(b.2) Additive fusion:** MAE embeddings and CNN features are aligned in shape and dimension, and then perform element-wise summation.
**(b.3) Attention fusion:** The CNN features are treated as query tokens and fused with MAE embeddings (key, value tokens) using a lightweight cross-attention module, allowing dynamic weighting of global context.

Same as the U-Net encoder-decoder structure, skip connections preserve resolution details from earlier stages, and upsampling layers progressively restore spatial dimensions. The final convolutional segmentation head outputs dense pixel-wise predictions for brain tissues and anatomical structures. Notably, the MAE weights remain frozen to ensure stable generalization under few-shot conditions while reducing the trainable parameter size. The training recipe is set to use the AdamW optimizer, a $1 \times 10^{-4}$ learning rate, and a batch size of 48.

### 2.3.2 Hybrid Loss Function

To achieve competitive performance on multiple brain regions segmentation, especially when handling class imbalance caused by the rarity of certain regions due to size and anatomical location, we introduce a composite loss function that integrates *Dice loss*, *Focal loss*, and *Cross-Entropy loss*. Each component targets different challenges in medical segmentation, including class imbalance, partial volume effects, and small-structure delineation. The final loss is a weighted sum of the three terms:

$$\mathcal{L}_{total} = \lambda_{dice} \cdot \mathcal{L}_{dice} + \lambda_{focal} \cdot \mathcal{L}_{focal} + \lambda_{ce} \cdot \mathcal{L}_{ce} \tag{3}$$

where $\lambda_{dice}$, $\lambda_{focal}$, $\lambda_{ce}$ are the weighting coefficients, all empirically set to 1.0 in our experiments. The detailed mathematics for each loss term is explained below.

**Dice Loss:** We use a non-batch variant of the soft Dice loss computed per sample, including the background class (unmasked area). Let $p_{i,c}$ and $g_{i,c}$ be the softmax prediction and the ground truth indicator for the pixel $i$ and class $c$, then:

$$\mathcal{L}_{dice} = 1 - \frac{2 \sum_i p_{i,c} g_{i,c} + \epsilon}{\sum_i p_{i,c}^2 + \sum_i g_{i,c}^2 + \epsilon} \tag{4}$$

where $\epsilon = 10^{-5}$ is a smoothing constant to prevent division by zero.

**Focal Loss:** To address class imbalance, we adopt the standard focal loss formulation:

$$\mathcal{L}_{focal} = -\alpha \left(1 - p_{i,c}\right)^\gamma \log\left(p_{i,c}\right) \tag{5}$$

where $\gamma$ = 2.0, $\alpha$ = 0.25, and $p_{i,c}$ is the predicted probability for the true class. This loss penalizes easy examples less and focuses training on hard misclassified pixels.

**Cross-Entropy Loss:** We include the standard pixel-wise multi-class cross-entropy loss:

$$\mathcal{L}_{ce} = -\sum_c g_{i,c} \log\left(p_{i,c}\right) \tag{6}$$

This term enforces accurate classification at the pixel level and complements the Dice and Focal components for better convergence.

### 2.3.3 Evaluation Metrics

To quantitatively evaluate the performance of our MAE-FUnet on segmentation tasks, we adopt two frequently used evaluation metrics in medical image segmentation: the Dice score and Intersection over Union (IoU). These two metrics are often used for evaluating performance on imbalanced and multi-class anatomical structures, as they provide meaningful overlap measures between predicted and ground-truth segmentations.

**Dice Score:** The Dice Score, also known as the F1 score in binary segmentation, is defined as:

$$Dice = 2 \times \frac{|P \cap G|}{|P| + |G|} \tag{7}$$

where $P$ is the set of predicted positive pixels and $G$ is the set of ground-truth positive pixels.

**Intersection over Union (IoU):** The Intersection over Union, also referred to as the Jaccard Index, is defined as:

$$IoU = \frac{|P \cap G|}{|P \cup G|} \tag{8}$$

where $P$ and $G$ are the same as in Dice score Equation (7).

### 2.4 Datasets

#### 2.4.1 National Alzheimer's Coordinating Center (NACC)

The NACC dataset, maintained by the National Alzheimer's Coordinating Center, is a large, publicly available database used extensively in neuroscience and medical research. The version used in this study is the latest Uniform Data Set (UDS): Version 3 (UDSv3) with the Neuropathology Data Set (NP) currently at Version 11. Overall, it includes over 54,000 patients and more than 200,000 clinical assessments, with MRI records spanning from the year 2005 to 2024. In our study, we filter out only the MRI sessions to construct a dataset with sequence modalities across T1, T2, FLAIR, DWI, DTI, and PD. The total number of 2D MRI slices from the NACC dataset is 8.5 million.

#### 2.4.2 Alzheimer's Disease Neuroimaging Initiative (ADNI)

Data used in the preparation of this article were obtained from the Alzheimer's Disease Neuroimaging Initiative (ADNI) database (adni.loni.usc.edu). The ADNI was launched in 2003 as a public-private partnership, led by Principal Investigator Michael W. Weiner, MD. The primary goal of ADNI has been to test whether serial magnetic resonance imaging (MRI), positron emission tomography (PET), other biological markers, and clinical and neuropsychological assessment can be combined to measure the progression of mild cognitive impairment (MCI) and early Alzheimer's disease (AD). The dataset used in this study is a combined cohort from all five versions: ADNI1, ADNI2, ADNI3, ADNI4, and ADNIDOD. The total dataset mainly focused on structured MRI, including T1, T2, FLAIR, and DWI, with a total number of 12 million 2D MRI slices.

#### 2.4.3 Open Access Series of Imaging Studies (OASIS)

OASIS is a public neuroimaging dataset designed to support research in aging, Alzheimer's disease, and neurodegeneration. The original release was developed by the Marcus Institute for Aging Research at Washington University. The dataset includes MRI scans, cognitive assessments, clinical metadata, and other longitudinal data. The MRI scans for our study are filtered from all 4 versions:

OASIS-1, OASIS-2, OASIS-3, and OASIS-4 (in progress), which provide a total of 10 million 2D MRI slices across all sequence modalities.

### 2.4.4 RadImageNet

RadImageNet is a large-scale radiology-specific image dataset created to support deep learning research in medical imaging, particularly in radiographic modalities like CT, MRI, and X-ray. Within this study, we pick only brain-related MRIs across categories of white matter changes, pituitary lesion, chronic infarction, and normal samples, etc. The entire dataset size of 2D MRI slices is 44,671.

### 2.4.5 FastMRI

FastMRI is a large-scale, open-source dataset and research initiative developed by Facebook AI Research (FAIR) in collaboration with NYU Langone Health, aimed at accelerating MRI reconstruction using deep learning. In our case, only brain-related multi-coil series are selected from sequences T1, T2, and FLAIR. The total number of MRI scans from the train, test, and validation subsets yields a total of 97,900 2D MRI slices.

### 2.4.6 Neurofeedback Skull-stripped (NFBS) Repository

The NFBS dataset is a repository of high-quality, manually skull-stripped T1-weighted anatomical MRI scans designed as a gold standard for brain extraction (skull-stripping) in neuroimaging research. As a part of the Enhanced Rockland Sample Neurofeedback Study, it contains T1-weighted MRI scans in NIfTI format from 125 participants, aged 21 to 45, with a variety of clinical and subclinical psychiatric symptoms. In our skull stripping experiment, we split it into a training dataset of randomly sampled 1 patient and a testing dataset of 62 patients.

### 2.4.7 SynthStrip

The SynthStrip dataset is an open-access collection designed for developing and evaluating skull-stripping (brain extraction) tools across a wide range of imaging modalities, populations, and protocols. It was originally created to test FreeSurfer software with 622 full-head 3D scans consisting of MRI, CT, and PET images. In this study, we randomly sampled 1 patient from each sequence (T1, T2, PD, FLAIR, and DWI) to form the training dataset, and a total of 311 MRI volumes for the testing dataset.

### 2.4.8 MRBrainS18

MRBrainS18 is part of the MR Brain Segmentation Challenge 2018, organized during the MICCAI 2018 conference. It is a public benchmarking dataset and competition platform designed to evaluate algorithms for multi-structured brain segmentation on 3T MRI scans, especially in the presence of large brain pathologies. In this study, we select only T1 MRI scans from 7 subjects to form the training dataset and 23 subjects to form the testing dataset. The anatomical segmentation focuses on 8 regions, excluding the hyperintensity class.

## 3. Results

If not explicitly indicated, all experiments follow the same training recipe with the AdamW optimizer, $1 \times 10^{-4}$ learning rate, and batch size of 48. The data preprocessing involves standard max-min normalization with 0.1%-99.9% clamping, followed by data augmentation using random flipping, rotation, and center cropping.

### 3.1 Sequence Detection

To evaluate the classification capability of pretrained MRI transformers, we deploy them onto one of the commonly studied medical classification tasks—sequence detection [62-65]. To tackle this task under few-shot conditions, we construct the training dataset comprising MRIs across seven MRI sequence types: T1, T2, FLAIR, PD, T2* [66], Susceptibility-Weighted Imaging (SWI) [67], and DTI/DWI. For each sequence, we randomly sample *n* slices from three different datasets: NACC, OASIS, and ADNI, respectively. By procedurally increasing *n* from 10 to 100, we simulate the few-shot conditions with differing dataset sizes. For comparison, we benchmark our method, *MAE-classify* (introduced in Section 2.2), against established architectures including U-Net, ResNet, EfficientNetv2, and MedViT. As shown in Table 1 and Figure 4a, our lightweight MAE-based classifier consistently achieves state-of-the-art accuracy across all individual modalities, with an overall accuracy of 99.24%, and outperforms all other methods.

| Method | Accuracy (%) | | | | | | | | Trainable Params |
|---|---|---|---|---|---|---|---|---|---|
| | T1 | T2 | FLAIR | PD | T2* | SWI | DTI/DWI | Total | |
| U-Net | 94.62 | 94.19 | 86.12 | 89.94 | 96.63 | 95.3 | 97.06 | 93.42 | 32M |
| EfficientNetV2 | 98.83 | 97.85 | 98.71 | 96.82 | 98.64 | 98.93 | 99.87 | 98.59 | 20M |
| ResNet | 98.09 | 98.39 | 97.17 | 95.66 | 98.61 | 98.34 | **99.90** | 98.10 | 11M |
| MedViT | 94.94 | 95.17 | 94.85 | 91.83 | 97.16 | 92.28 | 98.82 | 95.05 | 31M |
| *MAE-classify* | **99.51** | **99.43** | **99.64** | **97.39** | **99.42** | **99.31** | 99.51 | **99.24** | **6152** |

**Table 1.** MRI Sequence classification accuracy across modalities and models with their trainable parameters in millions (M). Accuracy (%) is reported for each MRI modality, with the training set consisting of 30 slices per sequence per dataset.

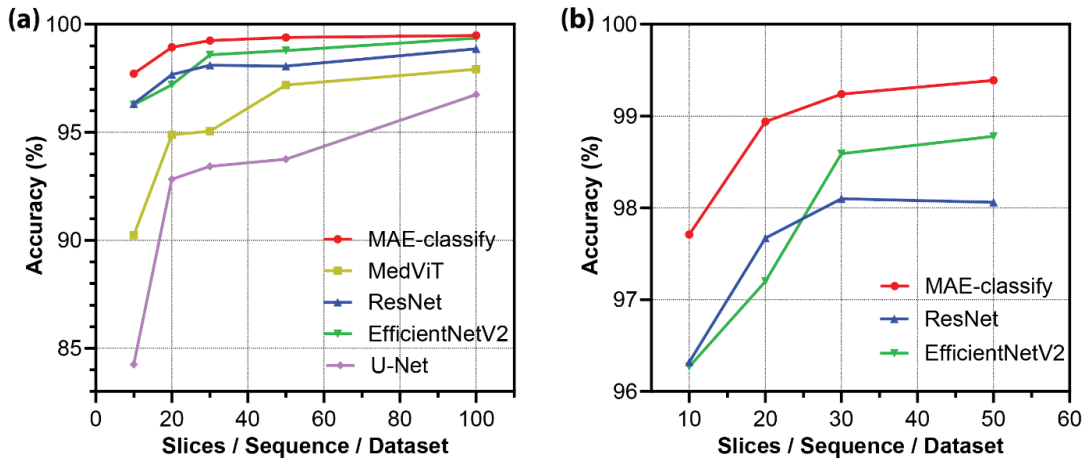

**Figure 4.** Performance of MRI sequence classification under varying sample sizes. (a) Classification accuracy across 7 MRI modalities with the number of slices increasing from 10 to 100. (b) Zoom in performance on MAE-classify, ResNet, and EfficientNetV2 with limited slices from 10 to 50.

Specifically, in Table 1 with 30 slices per sequence per dataset, MAE-classify achieved the highest classification accuracy in six out of seven categories, including T1 (99.51%), T2 (99.43%), T2 FLAIR (99.64%), PD (97.39%), *T2 Hemo (99.42%), and *T2 SWI (99.31%). Even in the DTI/DWI category, where ResNet slightly surpasses it with 99.90%, MAE-classify still maintains a strong performance at 99.51%. This result highlights the pretrained transformers' ability to encode global contextual features that are critical for distinguishing fine-grained sequence characteristics. Additionally, MAE-classify achieved this with only 6,152 trainable parameters, while other models often take up 11-32 million trainable parameters. Such a comparable size difference demonstrates the efficiency of self-supervised pretraining and minimal task-specific tuning, especially when encountering direct classification such as modality detection.

We also perform few-shot evaluations to further examine sample efficiency, testing the model with different amounts of input slices per sequence. As illustrated in Figure 4, although all models benefit from an increased amount of data, MAE-classify outperforms other models in all dataset sizes in terms of data efficiency. It approaches 98.9% accuracy with just 20 slices, while even the most competitive models, such as ResNet and EfficientNetv2, require significantly more samples to get comparable performance. Meanwhile, U-Net and MedViT converge slowly and tend to plateau at lower accuracies, even with 100 slices. These observations confirm the robustness of MAE-classify in few-shot scenarios, which is important in medical practice where annotated data is often limited.

### 3.2 Skull Stripping

To evaluate the segmentation performance of our proposed MAE-FUnet architecture on the task of brain skull stripping, we conduct comparisons across multiple model baselines and datasets. The models used for comparisons are state-of-the-art machine learning architectures, particularly designed for image segmentation. Each baseline is carefully chosen and falls into different architecture categories. U-Net is a pure CNN-based structure with the same architecture details as the U-Net backbone used in MAE-FUnet. Segformer is a pure transformer-based architecture with multi-scale skip connections. Swin-Unet is also transformer-based and constructed as a U-Net architecture, but with the Swin Transformer [68] replacing CNN modules. TransUNet is a hybrid architecture that combines both transformer and U-Net structures, with transformer modules appended only at the bottleneck layer of the U-Net. The MAE-direct structure is similar to the direct classification model in Figure 2, with a multi-layer CNN segmentation head replacing the classification head.

**Controlled Data Sampling via Stride:** We employ a controlled sampling technique with stride size to approximate the few-shot case with limited data. For each MRI sequence, we randomly select a single subject's volume and extract 2D slices along the acquisition axis. Note that if the MRI volume is 3D, we resize the volume by the voxel size fraction before extracting 2D slices along all three dimensions. Instead of using all available slices, we sample every *k-th* slice along the acquisition dimensions, where *k* is the stride used. This simulates real-world constraints where only sparse annotations or limited slices are available per patient, especially when MRI is acquired with large slicing thickness. For example, a stride of 5 results in doubling the number of training slices compared to a stride of 10, allowing us to regulate the size of the training data in a systematic manner. This strategy ensures anatomical diversity while maintaining consistency across models and runs.

| Method | NFBS | SynthStrip | | | | |
|---|---|---|---|---|---|---|
| | T1 | T1 | T2 | FLAIR | PD | DWI |
| U-Net | 92.96 / 96.35 | 91.38 / 95.23 | 94.00 / 96.89 | 91.67 / 95.65 | 93.99 / 96.87 | 88.49 / 93.82 |
| Segformer | 94.21 / 97.02 | 93.67 / 96.68 | 93.97 / 96.88 | 93.26 / 96.51 | 94.04 / 96.92 | 90.89 / 95.18 |
| TransUNet | 94.13 / 96.97 | 90.05 / 94.35 | 91.98 / 95.76 | 92.98 / 96.36 | 93.69 / 96.72 | 89.76 / 94.52 |
| Swin-Unet | 90.33 / 94.92 | 81.15 / 88.93 | 81.56 / 89.07 | 80.99 / 89.24 | 87.34 / 92.97 | 75.21 / 85.10 |
| MAE-direct | 90.25 / 94.88 | 91.70 / 95.62 | 92.85 / 96.28 | 92.86 / 96.30 | 92.96 / 96.34 | 88.56 / 93.88 |
| *MAE-FUnet* | **96.57 / 98.26** | **95.33 / 97.54** | **95.52 / 97.70** | **94.32 / 97.08** | **95.56 / 97.72** | **90.93 / 95.20** |

**Table 2.** Skull stripping performance (IoU / Dice%) across NFBS and SynthStrip datasets (sample stride = 7). Results are reported for five MRI sequences (T1, T2, FLAIR, PD, DWI) in SynthStrip and T1 in NFBS. Each value pair represents IoU / Dice.

In Table 2, we report both Dice and IoU scores on two representative brain datasets: NFBS and SynthStrip. The stride size is fixed to 7 for all datasets and sequence modalities. The proposed MAE-FUnet used in this experiment consists of four downsampling/upsampling layers plus a bottleneck layer, with the initial feature dimension set to 64. The corresponding MAE pretrained transformer layers selected for fusion are set to be the 1st, 3rd, 6th, 9th, and 12th layers from its encoder. The fusion strategy is set to concatenation. MAE-FUnet achieves the highest Dice and IoU scores in nearly all modalities on both datasets, indicating its superior adaptability and efficiency for latent space fusion between CNN features and transformer embeddings. Furthermore, the underperformance of MAE-direct compared to other baselines, along with the superior performance of MAE-FUnet over MAE-direct, validates the effectiveness and necessity of the proposed fusion mechanisms.

| Sample Stride | *MAE-FUnet* | MAE-direct | Swin-Unet | U-Net | TransUNet | Segformer |
|---|---|---|---|---|---|---|
| 4 | **95.16 / 97.52** | 92.37 / 96.03 | 81.69 / 89.89 | 93.03 / 96.38 | 92.11 / 95.89 | 94.52 / 97.18 |
| 5 | **95.24 / 97.56** | 92.54 / 96.12 | 83.50 / 90.99 | 93.06 / 96.40 | 92.46 / 96.08 | 94.49 / 97.16 |
| 6 | **95.23 / 97.55** | 92.11 / 95.89 | 83.87 / 91.21 | 93.07 / 96.41 | 92.41 / 96.05 | 94.28 / 97.05 |
| 7 | **95.25 / 97.57** | 92.41 / 96.05 | 82.06 / 90.12 | 92.96 / 96.35 | 91.55 / 95.58 | 93.86 / 96.83 |
| 8 | **95.18 / 97.53** | 92.40 / 96.05 | 82.58 / 90.44 | 92.72 / 96.22 | 92.23 / 95.96 | 93.98 / 96.90 |
| 9 | **95.16 / 97.52** | 92.45 / 96.08 | 83.61 / 91.05 | 92.77 / 96.25 | 92.05 / 95.86 | 93.85 / 96.83 |
| 10 | **95.12 / 97.50** | 92.17 / 95.92 | 77.76 / 87.45 | 92.59 / 96.15 | 92.31 / 96.00 | 93.76 / 96.78 |
| Mean | **95.19 / 97.54** | 92.35 / 96.02 | 82.15 / 90.16 | 92.89 / 96.31 | 92.16 / 95.92 | 94.11 / 96.96 |
| STD (%) | **0.045 / 0.023** | 0.142 / 0.077 | 1.949 / 1.200 | 0.176 / 0.094 | 0.284 / 0.156 | 0.295 / 0.154 |

**Table 3.** Evaluation on SynthStrip with varying sample strides. Lower stride values correspond to denser slice sampling (more training samples), while higher strides reduce the training set size. Mean and standard deviation (STD) of IoU and Dice scores are provided to assess performance stability under limited data.

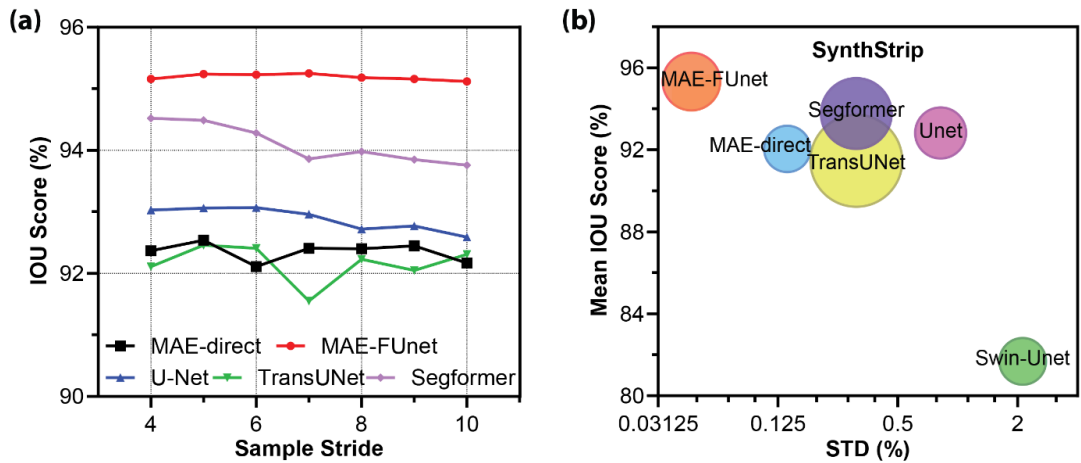

**Figure 5.** Performance of MRI skull stripping under varying sample sizes. (a) IoU score vs. sample stride on SynthStrip dataset. The line plot illustrates the segmentation performance trend as sample availability decreases. (b) Mean IoU vs. standard deviation (STD) for all models on SynthStrip. The bubble chart visualizes the trade-off between accuracy and robustness. Bubble size reflects trainable parameters across various baselines.

The robustness under sample scarcity is tested by varying the stride size from 4 to 10 on SynthStrip dataset, which offers comprehensive MRI modalities and a larger dataset size for validation (63,120 in SynthStrip vs 23,424 in NFBS). The performance comparison is reported in Table 3 as the mean Dice and IoU scores across all sequences and datasets. As expected, increasing the stride (i.e., reducing training data) leads to overall segmentation accuracy degradation. However, MAE-FUnet outperforms others in stability, showing exceptional robustness with a mean IoU of 95.19% and the lowest standard deviation of 0.045%. This is also observed visually in Figure 5, where MAE-FUnet not only always outperforms other models but also has the most stable trend line over increasing stride size. Figure 5b is a bubble graph that shows IoU accuracy vs. variance of all baselines, where the size of the bubble represents the number of trainable parameters. We can observe MAE-FUnet residing in the top-left quarter with a medium model size. This implies that MAE-FUnet achieves a high mean

IoU and low performance fluctuation while maintaining a relatively efficient parameter range. On the other hand, models like Swin-Unet show both lower mean and higher variance, resulting in less reliability in data-scarce conditions.

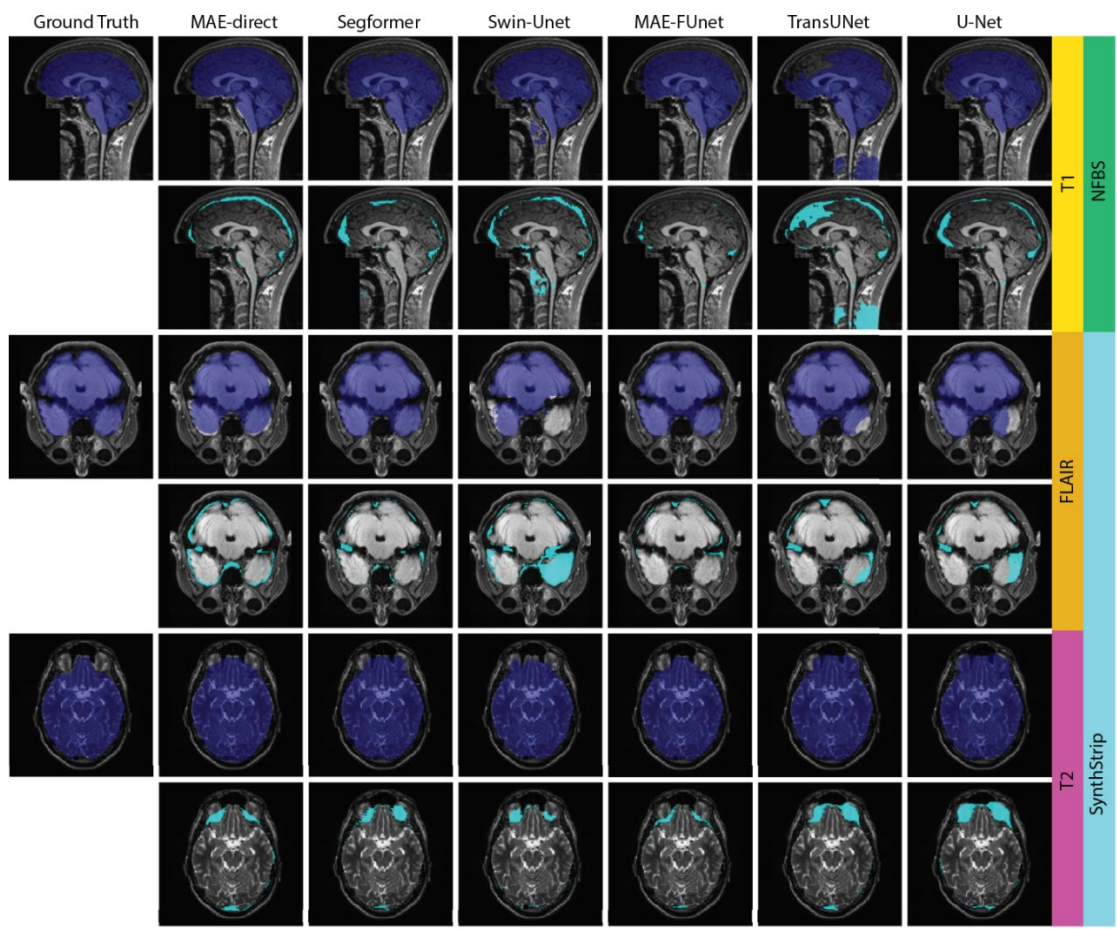

**Figure 6.** Visual comparison of skull stripping predictions across MRI sequences. Each row corresponds to a different MRI sequence (T1, T2, FLAIR for demonstration) from the NFBS and SynthStrip datasets. The top images show the predicted binary brain masks, and the bottom images display pixel-wise error maps (cyan) highlighting misclassified regions relative to ground truth.

We also include a qualitative visualization in Figure 6 to complement our quantitative results. It shows representative skull stripping outcomes across multiple sequences from both the NFBS and SynthStrip datasets. The comparison is done on both predicted brain masks (top of each row) and corresponding error maps (bottom of each row), where cyan highlights mislabeled pixels. These examples demonstrate that the MAE-FUnet achieves superior structure fidelity and minimal boundary error. On the other hand, some other baseline models struggle with both including non-brain regions within the brain mask like tissues with the cervical spine that are most notable with the TransUNet, or failing to include brain regions in the mask like the Swin-Unet failing to include portions of the anterior temporal lobes. In short, these results confirm that: (1) directly deploying pretrained MAE with appending modules, such as CNN heads, results in inferior segmentation, and (2) fusing hierarchical CNN features with MAE latent embeddings, as in MAE-FUnet, can substantially boost anatomical delineation, particularly in few-shot settings.

### 3.3 Multi-Class Anatomical Segmentation

To further evaluate the performance of MAE-FUnet on more complex segmentation tasks, we conduct experiments for multi-class anatomical segmentation using two benchmark datasets: MRBrainS18 and NACC. The MRBrainS18 dataset consists of 7 training samples and 23 testing samples, all hand-annotated with 8 brain regions. While the NACC dataset does not originally contain any annotations, we use FreeSurfer to generate sufficient segmentation masks to form training and testing datasets. For the NACC dataset, we choose 13 commonly observed brain regions as targets. All samples are restricted to T1 sequences, as such modality often yields the most accurate and stable results in both hand annotations and software-generated outputs. The model used for this experiment is identical to the one applied in skull stripping: MAE-FUnet with 4-layer depth, an initial feature dimension of 64, and a concatenation-based fusion strategy.

| Region | IoU / Dice (%) | | | | |
|---|---|---|---|---|---|
| | *MAE-FUnet* | Swin-Unet | U-Net | TransUNet | Segformer |
| | MRBrainS18 | | | | |
| Cortical gray matter | **71.64 / 83.48** | 66.90 / 80.17 | 71.46 / 83.35 | 70.76 / 82.88 | 68.71 / 81.45 |
| Basal ganglia | **68.81 / 81.48** | 58.48 / 73.76 | 66.09 / 79.54 | 67.29 / 80.34 | 62.86 / 77.12 |
| White matter | **76.56 / 86.72** | 71.84 / 83.61 | 74.95 / 85.68 | 75.21 / 85.84 | 73.17 / 84.50 |
| White matter lesions | 40.10 / 57.01 | 27.24 / 42.47 | 37.84 / 54.68 | **40.39 / 57.29** | 35.03 / 51.80 |
| Cerebrospinal fluid | **67.14 / 80.33** | 60.84 / 75.64 | 66.69 / 80.01 | 65.45 / 79.10 | 65.06 / 78.82 |
| Ventricles | **87.06 / 93.07** | 81.16 / 89.59 | 85.72 / 92.31 | 85.47 / 92.14 | 80.56 / 89.21 |
| Cerebellum | **87.16 / 93.14** | 76.10 / 86.40 | 86.23 / 92.60 | 85.17 / 91.58 | 84.63 / 91.67 |
| Brain stem | 80.80 / 89.27 | 73.25 / 84.53 | 78.17 / 87.62 | **81.90 / 89.89** | 77.03 / 86.89 |
| **Mean** | **72.41 / 83.06** | 64.48 / 77.02 | 70.89 / 81.97 | 71.46 / 82.38 | 68.38 / 80.18 |
| | NACC | | | | |
| Cerebral White Matter | **87.03 / 93.06** | 83.71 / 91.13 | 86.44 / 92.72 | 85.70 / 92.29 | 86.58 / 92.80 |
| Cerebral Cortex | **77.96 / 87.60** | 72.06 / 83.74 | 76.46 / 86.65 | 74.47 / 85.35 | 77.40 / 87.25 |
| Cerebellum White Matter | **75.91 / 86.11** | 67.46 / 80.36 | 73.27 / 84.46 | 71.88 / 83.47 | 73.89 / 84.84 |
| Cerebellum Cortex | **81.60 / 89.77** | 73.11 / 84.35 | 79.26 / 88.36 | 77.25 / 87.06 | 78.98 / 88.17 |
| Thalamus | **79.12 / 88.17** | 69.99 / 82.15 | 78.01 / 87.54 | 74.04 / 84.92 | 74.52 / 85.26 |
| Caudate | **75.04 / 85.54** | 63.80 / 77.62 | 70.51 / 82.51 | 68.83 / 81.30 | 67.24 / 80.15 |
| Putamen | **74.64 / 85.24** | 62.20 / 76.44 | 71.62 / 83.28 | 70.92 / 82.76 | 65.06 / 78.54 |
| Pallidum | **66.61 / 79.50** | 51.64 / 67.59 | 63.51 / 77.40 | 63.31 / 77.18 | 58.83 / 73.60 |
| Brainstem | **80.22 / 88.78** | 71.84 / 83.39 | 77.69 / 87.30 | 77.72 / 87.26 | 77.11 / 86.87 |
| Hippocampus | **73.24 / 84.30** | 54.83 / 70.48 | 66.68 / 79.82 | 64.68 / 78.25 | 62.09 / 76.30 |
| Amygdala | **64.79 / 78.14** | 40.83 / 57.24 | 58.27 / 73.24 | 57.51 / 72.56 | 52.73 / 68.45 |
| CSF | **57.71 / 72.40** | 28.20 / 42.70 | 52.45 / 68.19 | 53.74 / 69.26 | 34.59 / 50.48 |
| WM-hypointensities | **53.60 / 69.52** | 41.69 / 58.52 | 50.21 / 66.69 | 47.47 / 64.12 | 45.88 / 62.68 |
| **Mean** | **72.88 / 83.70** | 60.10 / 73.52 | 69.57 / 81.40 | 68.27 / 80.44 | 65.76 / 78.11 |

**Table 4.** Dice and IoU scores (%) for anatomical structure segmentation with fixed 30 training samples on both MRBrainS18 and NACC datasets. Each dataset contains evaluation for separate regions as well as overall Dice/IoU mean scores across all regions.

By training on a dataset of 30 samples for both MRBrainS18 and NACC, the results reported in Table 4 show that in both datasets, MAE-FUnet consistently outperforms baseline models, including U-Net,

Swin-Unet, TransUNet, and Segformer, across most brain regions. On MRBrainS18, it achieves the highest mean Dice score of 83.06% and IoU score of 72.41%, with especially strong performance on critical regions such as the cerebellum, ventricles, and brainstem. A similar trend is observed on NACC, where MAE-FUnet ranks first, with a mean Dice score of 83.70% and an IoU of 72.88%. Upon further observation, the MAE-FUnet also demonstrates exceptional accuracy in segmenting small and complex regions. For example, MAE-FUnet outperforms the second-best baseline by +4.48% in Dice score on the hippocampus and by +4.90% on the amygdala, while achieving more modest improvements of +0.26% and +0.56% in relatively large regions such as cerebral white matter and cerebral cortex.

| Sample Size | MAE-FUnet | Swin-Unet | U-Net | TransUNet | Segformer |
|---|---|---|---|---|---|
| 10 | **69.45 / 81.31** | 47.81 / 62.88 | 58.73 / 72.79 | 66.90 / 79.42 | 60.44 / 74.08 |
| 20 | **69.37 / 81.25** | 59.26 / 73.33 | 64.01 / 77.11 | 68.49 / 80.63 | 62.68 / 75.37 |
| 30 | **72.88 / 83.70** | 60.10 / 73.52 | 67.54 / 79.93 | 68.27 / 80.44 | 65.76 / 78.11 |
| 50 | **71.89 / 83.03** | 59.87 / 73.60 | 66.46 / 78.95 | 70.08 / 81.78 | 67.34 / 79.58 |
| 100 | **74.88 / 85.17** | 66.39 / 78.94 | 70.16 / 81.86 | 73.83 / 84.43 | 72.66 / 83.52 |
| Mean | **71.70 / 82.89** | 58.69 / 72.45 | 65.38 / 78.13 | 69.51 / 81.34 | 65.77 / 78.13 |
| STD (%) | **2.100 / 1.489** | 6.024 / 5.232 | 3.864 / 3.081 | 2.380 / 1.718 | 4.192 / 3.324 |

**Table 5.** Few-shot segmentation evaluation recorded as overall IoU and Dice score for all 13 regions on NACC. By varying sample size from 10 to 100, the mean and standard deviation (STD) of IoU and Dice scores are provided to assess performance stability under limited data.

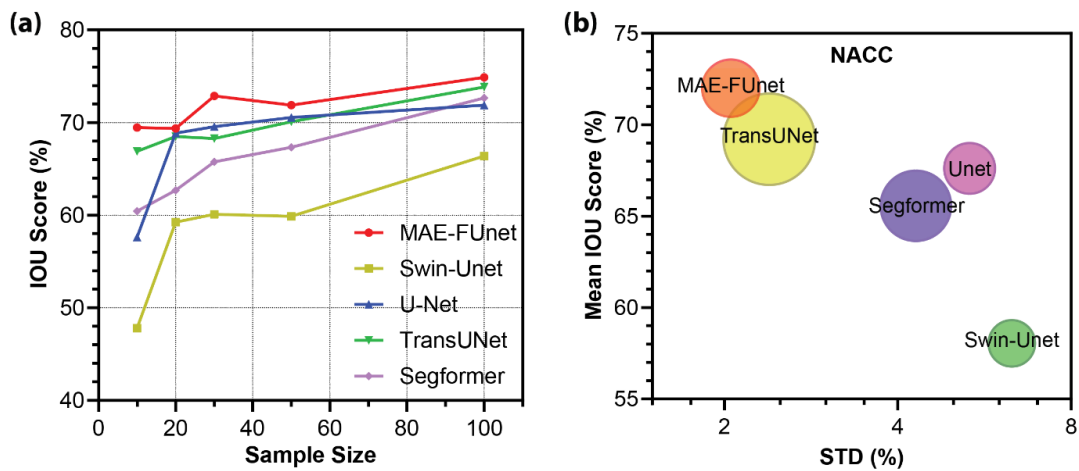

**Figure 7.** Performance of brain MRI multi-class segmentation under varying sample sizes. (a) IoU score trend across increasing sample sizes for few-shot anatomical segmentation on the NACC dataset. (b) Mean IoU vs. performance stability (STD) for all methods on the NACC dataset. Bubble size indicates the model's trainable parameter count.

Moreover, to further investigate robustness under data scarcity, as shown in Table 5, we simulate few-shot segmentation on the NACC dataset by reducing the number of training MRI volumes to 10, 20, 30, 50, and 100. As expected, performance improves via incrementing dataset size, but MAE-FUnet maintains its advantage across all sample sizes. Notably, it obtains an average Dice score of 82.89% and IoU of 71.70% across all volumes, which are superior to all other models by a large margin. It also has the lowest variance, with a standard deviation of 2.100% in IoU and 1.489% in Dice score, suggesting the best stability.

The trends are further visualized in Figure 7a and 7b. The line plot in Figure 7a highlights the consistently higher IoU scores at all sample sizes for MAE-FUnet. The bubble chart in Figure 7b illustrates the relationship between mean IoU and standard deviation, with the bubble size representing the number of trainable parameters for each model. Since MAE-FUnet occupies the very top-left quadrant, it obtains both the high mean IoU and low performance fluctuation. Considering the trainable parameter counts, these results demonstrate that MAE-FUnet achieves strong accuracy and robustness with a relatively modest model size.

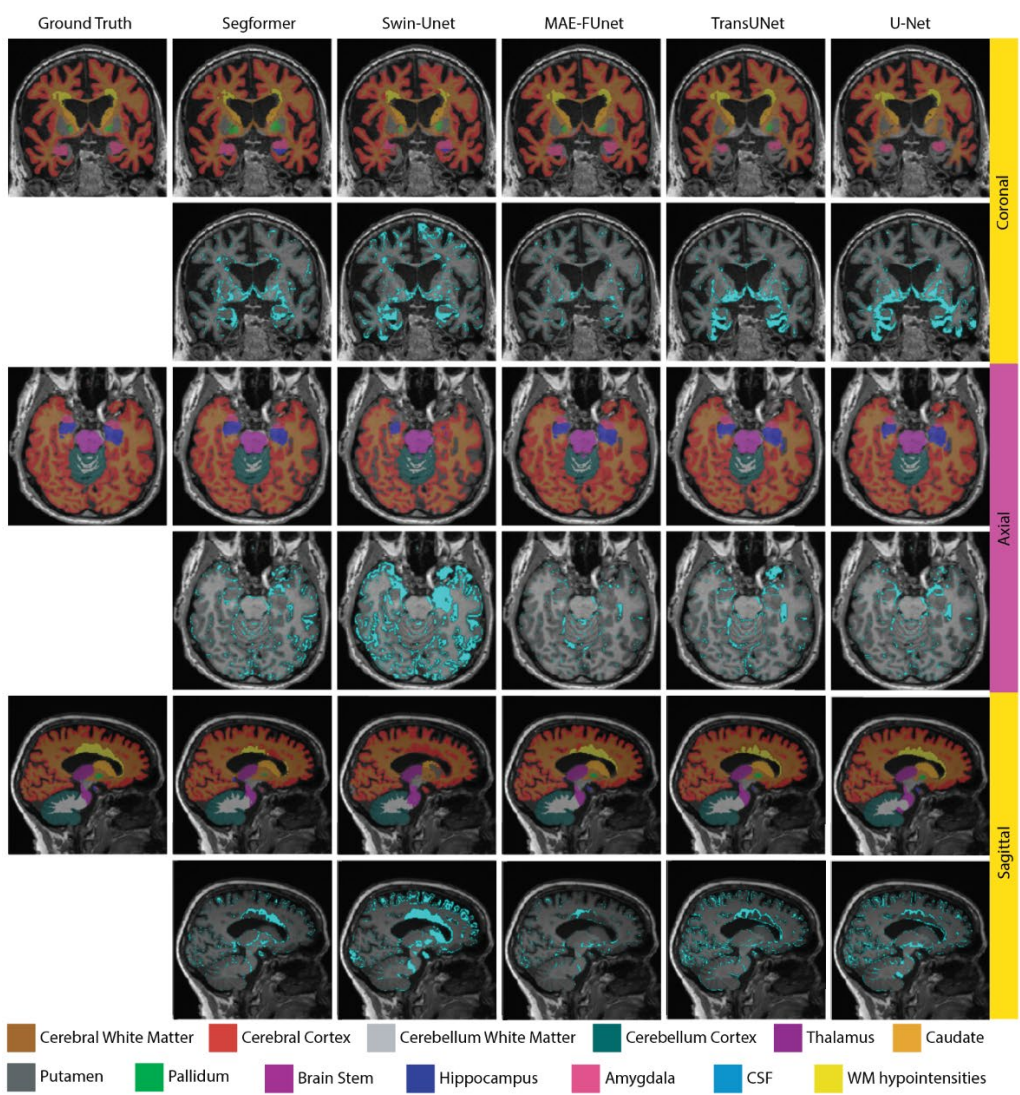

**Figure 8.** Qualitative segmentation results trained on 30 samples from the NACC dataset, demonstrated by Sagittal, Coronal, and Axial views on each row. For each row, the top includes predicted brain region masks, and the bottom includes error maps with cyan highlighting mislabeled pixels.

Qualitative comparisons across different models are provided in Figure 8, including sampled sagittal, coronal, and axial views of 13 segmenting regions in the NACC dataset. Each view includes segmentation predictions on the top row and corresponding error maps on the bottom row. In general, MAE-FUnet demonstrates exceptional anatomical accuracy with fewer boundary artifacts and better segmentation of complex regions. Similar to the results observed in the skull stripping task, MAE-FUnet tends to produce sharper, cleaner boundaries compared to other models, particularly in regions with narrow and elongated anatomical contours like the lateral ventricles. Furthermore, MAE-FUnet is superior in capturing small regions such as the hippocampus, amygdala, and putamen, while other models either under-segment (e.g., missing the amygdala) or over-segment (e.g., spilling over into neighboring tissue). The error maps further indicate MAE-FUnet's advantage with much fewer mislabeled pixels, represented by reduced cyan overlays compared to other baselines. In summary, qualitative and quantitative results together demonstrate MAE-FUnet's higher segmentation fidelity, especially for complex or small anatomic regions. By fusing CNN hierarchical features with learned MAE embeddings, the proposed model is able to preserve both local detail and global consistency in few-shot learning, while other baselines tend to struggle.

## 4. Ablation Study

To further investigate the contribution of key design components in MAE-FUnet, we perform ablation experiments focusing on two major aspects: (1) the impact of model sizes controlled by initial feature dimension, and (2) the choice of three fusion strategies, including addition, concatenation, and attention.

### 4.1 Effect of Feature Dimension Size

| Method | IoU / Dice (%) | | | | Trainable Params |
|---|---|---|---|---|---|
| | Skull Strip | | Multi-Class Segmentation | | |
| | NFBS | SynthStrip | MRBrainS18 | NACC | |
| MAE-FUnet32 | 96.15 / 98.04 | 95.21 / 97.55 | 71.27 / 82.19 | 70.62 / 82.08 | 10M |
| MAE-FUnet64 | **96.57 / 98.26** | **95.25 / 97.57** | 72.41 / 83.06 | 72.88 / 83.70 | 38M |
| MAE-FUnet96 | 96.27 / 98.10 | 95.19 / 97.53 | **73.21 / 83.64** | **73.27 / 83.85** | 84M |

**Table 6.** Ablation study on different model sizes controlled by initial feature dimension in MAE-FUnet across skull stripping and multi-class segmentation tasks. The performance is recorded by overall IoU and Dice loss on each dataset, along with the number of trainable parameters in millions (M).

In Table 6, we evaluate three MAE-FUnet variants: MAE-FUnet32, MAE-FUnet64, and MAE-FUnet96, corresponding to their initial convolutional channel dimensions of 32, 64, and 96, respectively. All other architectural and training settings remain unchanged as in Sections 3.2 and 3.3. The dataset size set for training is stride 7 for both NFBS and SynthStrip, 7 training samples for MRBrainS18, and 30 samples for NACC. By comparing all benchmarks, MAE-FUnet64 with its intermediate configuration achieves the best performance in NFBS and SynthStrip for skull stripping, while MAE-FUnet96 obtains the highest Dice and IoU scores in MRBrainS18 and NACC for multi-class segmentation.

It is notable that increasing feature dimension from 32 to 64 leads to improvements in both skull stripping and multi-class segmentation. Such a trend suggests that richer low-level feature representations are crucial for capturing anatomical context. However, further increasing the dimension to 96 does not lead to consistent improvements and may cause marginal performance drops for easier tasks, such as skull stripping. For instance, the IoU gain from MAE-FUnet64 to MAE-FUnet96 decreases by 0.3% on NFBS, despite more than doubling the parameter count from 38M to 84M. On the other hand, for more complex anatomical segmentation tasks presented in MRBrainS18 and NACC, all MAE-FUnet variants still have the potential of increasing performance with larger model sizes. These results are an important observation for few-shot learning: *increasing model complexity does not universally translate to better performance.* In fact, the optimal configuration can be influenced by the complexity of the target task, and balanced model design is required in data-limited conditions.

### 4.2 Fusion Strategy Analysis

| Method | IoU / Dice (%) | | | | Trainable Params |
|---|---|---|---|---|---|
| | **Skull Strip** | | **Multi-Class Segmentation** | | |
| | NFBS | SynthStrip | MRBrainS18 | NACC | |
| MAE-FUnet-add | 95.73 / 97.82 | 95.20 / 97.54 | 71.62 / 82.52 | 72.35 / 83.36 | 35M |
| MAE-FUnet-concat | **96.57 / 98.26** | **95.25 / 97.57** | **72.41 / 83.06** | **72.88 / 83.70** | 38M |
| MAE-FUnet-attent | 95.99 / 97.96 | 95.23 / 97.56 | 71.60 / 82.51 | 71.59 / 82.83 | 41M |

**Table 7.** Ablation study on fusion strategies—addition, concatenation, and attention—in MAE-FUnet across skull stripping and multi-class segmentation tasks. The performance is recorded by overall IoU and Dice loss on each dataset, along with the number of trainable parameters in millions (M).

In Table 7, we investigate different fusion mechanisms used to integrate MAE transformer embeddings into the CNN features. Specifically, we compare three strategies: additive fusion (MAE-FUnet-add), concatenation fusion (MAE-FUnet-concat), and attention-based fusion (MAE-FUnet-attent). All other architectural and training settings remain the same as in Sections 3.2 and 3.3, with the initial convolutional channel dimension fixed at 64. Among these three strategies, the concatenation approach achieves the best performance by attaining the highest Dice and IoU scores on all four datasets in both tasks, while maintaining a medium parameter size (38M). On the other hand, MAE-FUnet-attent provides slight improvements in some cases over the additive strategy, but is still not as competitive as MAE-FUnet-concat. In fact, attention-based strategy adds more computational overhead without consistent performance gains. Our results support that concatenation is the most effective fusion strategy, preserving and combining richer complementary information between CNN features and transformer embeddings. Therefore, this strategy leads to better spatial integration and segmentation accuracy, especially in the case of few-shot learning.

### 5. Conclusion

This study presents a systematic investigation into the few-shot deployment of pretrained MRI transformers for diverse, non-pathological brain imaging tasks, including both high-level tasks such as classification and low-level tasks such as segmentation. With pretraining workflow such as Masked Autoencoders (MAE), we show that it is feasible to obtain highly transferable latent representations from a large-scale, multi-cohort, unlabeled brain MRI corpus consisting of over 31 million 2D slices. These representations can be effectively re-utilized for multiple downstream tasks even in data-scarce conditions, which are common in clinical neuroimaging settings. Our study shows that pretrained MAE transformers offer strong generalizability across tasks, modalities, and datasets without the need for extensive data preprocessing (e.g., MRI registration, bias field correction, and reorientation) [69] or convoluted fine-tuning procedures. Thus, it is particularly suitable for fast deployments in low-resource medical imaging scenarios.

For classification tasks such as MRI sequence identification, we find that a simple linear classifier appended to a frozen MAE encoder is sufficient to outperform some well-established architectures like ResNet and EfficientNetv2. This highlights the strength of the learned MAE representations and their ability to capture semantic content with minimal task-specific adaptation. Through simulation on limited training data, the classification model achieves state-of-the-art performance at multiple few-shot conditions. Thus, it demonstrates the data efficiency and scalability of the proposed framework. Additionally, the extremely small number of trainable parameters in the classification head indicates the lightweight nature and the potential for fast deployment of MAE-based models in practice.

For segmentation, we introduce MAE-FUnet, a hybrid architecture that fuses multiscale CNN features with transformer embeddings from a pretrained MAE encoder, offering easy deployment and adjustment. Extensive tests on skull stripping and multi-class anatomical segmentation tasks over the NFBS, SynthStrip, MRBrainS18, and NACC datasets demonstrate that MAE-FUnet yields competitive accuracy, better few-shot robustness, and lower variance compared to other segmentation baselines such as TransUnet and Segformer. Ablation studies have also demonstrated that optimal performance relies on the feature dimension, fusion strategy, and carefully weighing the nature of task complexity. These results suggest that suitable architectural design is essential when integrating pretrained transformers with other machine learning modules to achieve desired performance, especially under data-scarce conditions.

More broadly speaking, our findings emphasize the importance of adjusting frameworks to fully invoke the potential of learned MAE embeddings, along with putting control on adaptation simplicity and model complexity. For example, it is proven effective to integrate latent space embedding from transformers into other module modalities (such as CNN) and achieve the best performance in medical image segmentation. Although CNNs are known to be efficient at capturing local imaging patterns, they often have difficulty learning global spatial coherence. Meanwhile, MAE-derived embeddings could potentially provide a rich, hierarchical understanding of the image at various scales. By combining both modules, our MAE-FUnet framework aims to successfully invoke the strengths of both convolutional and transformer-based models. This hybrid approach will be especially powerful in medical imaging, where fine-grained anatomical structures need to be accurately identified and where high-level spatial reasoning is also crucial. Furthermore, the adaptability and scalability of the MAE-FUnet provide more fine-tuning flexibility for various downstream tasks of differing natures. The framework supports customizable fusion strategies and tunable feature dimensionality, which enables task-specific optimization in various few-shot learning cases.

In conclusion, our work not only provides a practical, parameter-efficient pathway for few-shot medical AI with pretrained transformers but also gives a generalizable and reproducible fusion-based architecture design pipeline. Moving forward, we expect that our framework can be expanded to multi-modal brain imaging (e.g., MRI+PET), cross-domain generalization (utilization in other body locations), and downstream clinical applications, such as pathology detection and disease progression prediction [70, 71]. While transformer architecture continues to evolve and dataset continues to accumulate, this work lays the foundation for creating scalable, data-efficient, clinically deployable AI systems in medical imaging.

**Acknowledgements**

This work was supported by the Rajen Kilachand Fund for Integrated Life Science and Engineering and the Hariri Institute for Computing's Focused Research Programs (FRPs). We would like to thank the Boston University Photonics Center for technical support.

**NACC**
The NACC database is funded by NIA/NIH Grant U24 AG072122. NACC data are contributed by the NIA-funded ADRCs: P30 AG062429 (PI James Brewer, MD, PhD), P30 AG066468 (PI Oscar Lopez, MD), P30 AG062421 (PI Bradley Hyman, MD, PhD), P30 AG066509 (PI Thomas Grabowski, MD), P30 AG066514 (PI Mary Sano, PhD), P30 AG066530 (PI Helena Chui, MD), P30 AG066507 (PI Marilyn Albert, PhD), P30 AG066444 (PI David Holtzman, MD), P30 AG066518 (PI Lisa Silbert, MD, MCR), P30 AG066512 (PI Thomas Wisniewski, MD), P30 AG066462 (PI Scott Small, MD), P30 AG072979 (PI David Wolk, MD), P30 AG072972 (PI Charles DeCarli, MD), P30 AG072976 (PI Andrew Saykin, PsyD), P30 AG072975 (PI Julie A. Schneider, MD, MS), P30 AG072978 (PI Ann McKee, MD), P30 AG072977 (PI Robert Vassar, PhD), P30 AG066519 (PI Frank LaFerla, PhD), P30 AG062677 (PI Ronald Petersen, MD, PhD), P30 AG079280 (PI Jessica Langbaum, PhD), P30 AG062422 (PI Gil Rabinovici, MD), P30 AG066511 (PI Allan Levey, MD, PhD), P30 AG072946 (PI Linda Van Eldik, PhD), P30 AG062715 (PI Sanjay Asthana, MD, FRCP), P30 AG072973 (PI Russell Swerdlow, MD), P30 AG066506 (PI Glenn Smith, PhD, ABPP), P30 AG066508 (PI Stephen Strittmatter, MD, PhD), P30 AG066515 (PI Victor Henderson, MD, MS), P30 AG072947 (PI Suzanne Craft, PhD), P30 AG072931 (PI Henry Paulson, MD, PhD), P30 AG066546 (PI Sudha Seshadri, MD), P30 AG086401 (PI Erik Roberson, MD, PhD), P30 AG086404 (PI Gary Rosenberg, MD), P20 AG068082 (PI Angela Jefferson, PhD), P30 AG072958 (PI Heather Whitson, MD), P30 AG072959 (PI James Leverenz, MD).

**ADNI**
Data collection and sharing for this project was funded by the Alzheimer's Disease Neuroimaging Initiative (ADNI) (National Institutes of Health Grant U01 AG024904) and DOD ADNI (Department of Defense award number W81XWH-12-2-0012). ADNI is funded by the National Institute on Aging, the National Institute of Biomedical Imaging and Bioengineering, and through generous contributions from the following: AbbVie, Alzheimer's Association; Alzheimer's Drug Discovery Foundation; Araclon Biotech; BioClinica, Inc.; Biogen; Bristol-Myers Squibb Company; CereSpir, Inc.; Cogstate;

Eisai Inc.; Elan Pharmaceuticals, Inc.; Eli Lilly and Company; EuroImmun; F. Hoffmann-La Roche Ltd and its affiliated company Genentech, Inc.; Fujirebio; GE Healthcare; IXICO Ltd.; Janssen Alzheimer Immunotherapy Research & Development, LLC.; Johnson & Johnson Pharmaceutical Research & Development LLC.; Lumosity; Lundbeck; Merck & Co., Inc.; Meso Scale Diagnostics, LLC.; NeuroRx Research; Neurotrack Technologies; Novartis Pharmaceuticals Corporation; Pfizer Inc.; Piramal Imaging; Servier; Takeda Pharmaceutical Company; and Transition Therapeutics. The Canadian Institutes of Health Research is providing funds to support ADNI clinical sites in Canada. Private sector contributions are facilitated by the Foundation for the National Institutes of Health ([www.fnih.org] (http://www.fnih.org)). The grantee organization is the Northern California Institute for Research and Education, and the study is coordinated by the Alzheimer's Therapeutic Research Institute at the University of Southern California. ADNI data are disseminated by the Laboratory for Neuro Imaging at the University of Southern California.

**OASIS**

Data in this study related to the OASIS repository were provided by OASIS 1-4:

OASIS-1: Cross-Sectional: Principal Investigators: D. Marcus, R, Buckner, J, Csernansky J. Morris; P50 AG05681, P01 AG03991, P01 AG026276, R01 AG021910, P20 MH071616, U24 RR021382

OASIS-2: Longitudinal: Principal Investigators: D. Marcus, R, Buckner, J. Csernansky, J. Morris; P50 AG05681, P01 AG03991, P01 AG026276, R01 AG021910, P20 MH071616, U24 RR021382

OASIS-3: Longitudinal Multimodal Neuroimaging: Principal Investigators: T. Benzinger, D. Marcus, J. Morris; NIH P30 AG066444, P50 AG00561, P30 NS09857781, P01 AG026276, P01 AG003991, R01 AG043434, UL1 TR000448, R01 EB009352. AV-45 doses were provided by Avid Radiopharmaceuticals, a wholly owned subsidiary of Eli Lilly.

OASIS-3_AV1451: Principal Investigators: T. Benzinger, J. Morris; NIH P30 AG066444, AW00006993. AV-1451 doses were provided by Avid Radiopharmaceuticals, a wholly owned subsidiary of Eli Lilly.

OASIS-4: Clinical Cohort: Principal Investigators: T. Benzinger, L. Koenig, P. LaMontagne